\RequirePackage{fix-cm}
\documentclass[smallcondensed]{svjour3}       % 
\smartqed  % flush right qed marks, e.g. at end of proof
\usepackage{graphicx}
\usepackage{algorithm}
\usepackage{amssymb,array}
\usepackage{algcompatible}
\usepackage{amsmath}
\usepackage[figuresright]{rotating}
\usepackage{caption}
\usepackage{booktabs}
\usepackage{eqparbox}
\usepackage{setspace}
\usepackage{fancyhdr}
\usepackage{latexsym}
\usepackage[greek,english]{babel}
\usepackage{multirow}
\usepackage{calc}
\usepackage{tabulary}
\usepackage{csquotes}
\usepackage{bbm}
\usepackage{dsfont}
\usepackage{fixfoot}
\usepackage{scrextend}
\usepackage[usenames, dvipsnames]{color}

\usepackage{soul,color}

\usepackage{eqparbox}

\begin{document}

\title{Bootstrapping the Out-of-sample Predictions for 
Efficient and Accurate Cross-Validation}
\titlerunning{Efficient and Accurate Cross-Validation} 

\DeclareFixedFootnote{\repnote}{ This is a repeated footnote}

\author{Ioannis Tsamardinos\footnote{\label{first}Equal contribution.}\and
        Elissavet Greasidou\footref{first}\and
        Michalis Tsagris\and
        Giorgos Borboudakis
}

\authorrunning{Tsamardinos, I., Greasidou, E. et al.} % if too long for running head

\institute{Ioannis Tsamardinos \at
              Computer Science Department, University of Crete and Gnosis Data Analysis PC\\
              \email{tsamard.it@gmail.com} \and
              Elissavet Greasidou \at Computer Science Department, University of Crete and Gnosis Data Analysis PC \\
                  \email{greasidouelissavet@gmail.com}              
               \and
           Michalis Tsagris\at
              Computer Science Department, University of Crete
              \and
		Giorgos Borboudakis\at Computer Science Department, University of Crete and Gnosis Data Analysis PC\\
}

\date{Received: date / Accepted: date}

\maketitle

\begin{abstract}
Cross-Validation (CV), and out-of-sample performance-estimation protocols in general, are often employed both for (a) selecting the optimal combination of algorithms and values of hyper-parameters (called a configuration) for producing the final predictive model, and (b) estimating the predictive performance of the final model. However, the cross-validated performance of the best configuration is optimistically biased. We present an efficient bootstrap method that corrects for the bias, called Bootstrap Bias Corrected CV (BBC-CV). BBC-CV's main idea is to bootstrap the whole process of selecting the best-performing configuration on the out-of-sample predictions of each configuration, without additional training of models. In comparison to the alternatives, namely the nested cross-validation \cite{varma2006} and a method by Tibshirani and Tibshirani \cite{tibshirani2009}, BBC-CV is computationally more efficient, has smaller variance and bias, and is applicable to any metric of performance (accuracy, AUC, concordance index, mean squared error). Subsequently, we employ again the idea of bootstrapping the out-of-sample predictions to speed up the CV process. Specifically, using a bootstrap-based hypothesis test we stop training of models on new folds of statistically-significantly inferior configurations. We name the method Bootstrap Corrected with Early Dropping CV (BCED-CV) that is both efficient and provides accurate performance estimates.


\end{abstract}

\section{Introduction}

Typically, the goals of a machine learning predictive modelling task are two: to return a high-performing predictive model for operational use and an estimate of its performance. The process often involves the following steps: (a) {\em Tuning}, where different combinations of algorithms and their hyper-parameter values (called {\em configurations}) are tried producing several models, their performance is estimated, and the best configuration is determined, (b) {\em Production} of the final model trained on all available data using the best configuration, and (c) {\em Performance Estimation} of the final model.

Focusing first on tuning, we note that a configuration may involve combining several algorithms for every step of the learning process such as: pre-processing, transformation of variables, imputation of missing values, feature selection, and modeling. Except for rare cases, each of these algorithms accepts a number of hyper-parameters that tune its behavior. Usually, these hyper-parameters affect the sensitivity of the algorithms to detecting patterns, the bias-variance trade-off, the trade-off between model complexity and fitting of the data, or may trade-off computational complexity for optimality of fitting. Examples include the maximum number of features to select in a feature selection algorithm, or the type of kernel to use in Support Vector Machine and Gaussian Process learning. 

There exist several strategies guiding the order in which the different configurations are tried, from sophisticated ones such as Sequential Bayesian Optimization \cite{snoek2012,garnett2010} to simple grid search in the space of hyper-parameter values. However, independently of the order of production of configurations, the analyst needs to estimate the performance of the average model produced by each configuration on the given task and select the best.

The estimation methods of choice for most analysts are the out-of-sample estimation protocols, where a portion of the data training instances is hidden from the training algorithm to serve as an independent test set. The performance of several models stemming from different configurations is tried on the test set, also called the hold-out set, in order to select the best performing one. This procedure is known as the \textit{Hold-out} protocol. We will refer to such a test set as a {\em tuning set} to emphasize the fact that it is employed repeatedly by all configurations for the purposes of tuning the algorithms and the hyper-parameter values of the learning pipeline. We note that while there exist approaches that do not employ out-of-sample estimation, such as using the Akaike Information Criterion (AIC) \cite{akaike1974} of the models, the Bayesian Information Criterion (BIC) \cite{schwarz1978}, and others, in this paper we focus only on out-of-sample estimation protocols.

The process of withholding a tuning set can be repeated multiple times leading to several analysis protocols variations. The simplest one is to repeatedly withhold different, randomly-chosen tuning sets and select the 
one with the best average performance over all tuning sets. This protocol is called the \textit{Repeated Hold-out}.

Arguably however, the most common protocol for performance estimation for relatively low sample sizes is the {\em K-fold Cross-Validation} or simply Cross-Validation (CV). In CV the data training instances are partitioned to $K$ approximately equal-sized subsets, each one serving as a tuning set and the remaining ones as training sets. The performance of each configuration is averaged over all tuning folds. The difference with the Repeated Hold-Out is that the process is repeated exactly $K$ times and the tuning sets are enforced to be non-overlapping in samples (also referred to as instances or examples). The process can be repeated with different partitions of the data to folds leading to the {\em Repeated CV}.

A final note on tuning regards its name. In statistics, the term {\em model selection} is preferred for similar purposes. The reason is that a common practice in statistical analysis is to produce several models using different configurations on all of the available data, manually examine their fitting, degrees of freedom, residuals, AIC and other metrics and then make an informed (but manual) choice of a model. In contrast, in our experience machine learning analysts estimate the performance of each configuration and select the best {\em configuration} to employ on all data, rather than selecting the best model. Thus, in our opinion, the term tuning is more appropriate than model selection for the latter approach.

Considering now the production of the final model to deploy operationally, the most reasonable choice is arguably to {\em train a single model using the best configuration found on all of the available data}. Note that each configuration may have produced several models during CV or Repeated Hold-Out for tuning purposes, each time using a subset of the data for training. However, assuming that --- on average --- a configuration produces better predictive models when trained with larger sample sizes (i.e., its learning curve is monotonically increasing) it is reasonable to employ all data to train the final model and not waste any training examples (samples) for tuning or performance estimation. There may be exceptions of learning algorithms that do not abide to this assumption (K-NN algorithm for example, see \cite{krueger2015} for a discussion) but it is largely accepted and true for most predictive modeling algorithms.

The third step is to compute and return a performance estimate of the final model. {\em The cross-validated performance of the best configuration (estimated on the tuning sets) is an optimistically biased estimate of the performance of the final model. Thus, it should not be reported as the performance estimate.} Particularly for small sample sizes (less than a few hundred) like the ones that are typical in molecular biology and other life sciences, and when numerous configurations are tried, the optimism could be significant. This is in our opinion {\em a common source of methodological errors in data analysis}.

The main problem of using the estimation provided on the tuning sets is that these sets have been employed repeatedly by all configurations, out of which the analysts selected the best. Thus, equivalent statistical phenomena occur as in multiple hypothesis testing. The problem was named the multiple comparisons in induction problems and was first reported in the machine learning literature by Jensen in \cite{jensen2000}. A simple mathematical proof of the bias is as follows. Let $\mu_i$ be the average true performance (loss) of the models produced by configuration $i$ when trained on data of size $|D_\mathit{train}|$ from the given data distribution, where $|D_\mathit{train}|$ is the size of the training sets. The sample estimate of $\mu_i$ on the tuning sets is $m_i$, and so we expect that $\mu_i = E(m_i)$ for estimations that are unbiased. Returning the estimate of the configuration with the smallest loss returns $\min \{m_1, \ldots, m_n\}$, where $n$ is the number of configurations tried. On average, the estimate on the best configuration on the tuning sets is $E(\min \{m_1, \ldots, m_n\})$ while the estimate of the true best is $\min \{\mu_1, \ldots, \mu_n\} =  \min \{E(m_1), \ldots, E(m_n)\}$. The optimism (bias) is $\mathit{Bias} = \min \{E(m_1), \ldots, E(m_n)\} - E(\min \{m_1, \ldots, m_n\}) \geq 0$ by Jensen's inequality \cite{jensen1906}. %A straightforward proof of this is by simply noting that $E(\min \{m_1, \ldots, m_n\}) \leq E(m_i) \ \forall i=1,\ldots,n$.
For metrics such as classification accuracy and Area Under the Receiver's Operating Characteristic Curve (AUC) \cite{fawcett2006introduction}, where higher is better, the min is substituted with max and the inequality is reversed.

The bias of Cross-Validation when multiple configurations are tried has also been explored empirically in \cite{amin2014} on real datasets. For small samples ($<$ 100) the AUC bias ranges frequently between 5\% and 10\%. The bias depends on several factors such as (a) the number of configurations tried, (b) the correlation between the performance of the models trained by each configuration, (c) the sample size, and (d) the difference between the performance of the true best configuration and the rest.

To avoid this bias {\em the simplest procedure is to hold-out a second, untainted set, exclusively used for estimation purposes on a single model}. This single model is of course the one produced with the best configuration as found on the tuning sets. This approach has been particularly advocated in the Artificial Neural Networks literature were the performance of the current network is estimated on a {\em validation} set (equivalent to a tuning set) as a stopping criterion of training (weight updates). Thus, the validation set is employed repeatedly on different networks (models), albeit only slightly different by the weight updates of one epoch. For the final performance estimation a separate, independent test set is used. Thus, in essence, the data are partitioned to train-tuning-estimation subsets: the tuning set is employed multiple times for tuning of the configuration; then, a single model produced on the union of the train and tuning data with the best configuration is tested on the estimation subset. Generalizing this protocol so that all folds serve as tuning and estimation subsets and performance is averaged on all subsets leads to the {\em Nested Cross Validation} (NCV) protocol \cite{varma2006}. The problem with NCV is that it requires $O(K^2\cdot C)$ models to be trained, where $K$ the number of folds and $C$ the number of configurations tried, resulting in large computational overheads.

The main contribution of this paper is the idea that {\em one can bootstrap the pooled predictions of all configurations over all tuning sets (out-of-sample predictions)} to achieve several goals. The first goal is to estimate the loss of the best configuration (i.e., remove the $\mathit{bias}$ of cross validating multiple configurations) without training additional models. Specifically, the (out-of-sample) predictions of all configurations are bootstrapped, i.e., selected with replacement, leading to a matrix of predictions. The configuration with the minimum loss is selected and its loss is computed on the out-samples (not selected by the bootstrap). The procedure is repeated for a few hundred bootstraps and the average loss of the selected best configuration on the out-samples is returned. Essentially, the above procedure bootstraps the strategy for selecting the best configuration and computes its average loss on the samples not selected by the bootstrap.

Bootstrapping has a relatively low computational overhead and is trivially parallelized. The computational overhead for each bootstrap iteration amounts to re-sampling the sample indexes of predictions, computing the loss on the bootstrapped predictions for each configuration, and selecting the minimum. 
We call the method Bootstrap Bias Corrected CV (BBC-CV). BBC-CV is empirically compared against NCV, the standard for avoiding bias, and a method by Tibshirani and Tibshirani \cite{tibshirani2009} (TT from hereon) which addresses the large computational cost of NCV. BBC-CV is shown to exhibit a more accurate estimation of the $\mathit{Bias}$ than TT and similar to that of NCV, while it requires no training of new models, and thus being as computationally efficient as TT and much faster than NCV.
Bootstrapping the out-of-sample predictions can also trivially be used to compute confidence intervals for the performance estimate in addition to point estimates. 
In experiments on real data, we show that the confidence intervals are accurate or somewhat conservative (i.e. have higher coverage than expected).

The second main use of bootstrapping the out-of-sample predictions is to create a hypothesis test for the hypothesis that a configuration exhibits equal performance as the currently best configuration. The test is employed in every new fold serving for tuning during CV. When the hypothesis can be rejected based on the predictions on a limited number of folds, the configuration is eliminated or dropped from further consideration and no additional models are trained on remaining folds. We combine the idea of dropping configurations with the BBC-CV method for bias correction, and get the Bootstrap Corrected with Early Dropping CV (BCED-CV). BCED-CV results in significant computational gains, typically achieving a speed-up of 2-5 (in some cases up to the theoretical maximum equal to the number of folds, in this case 10) over BBC-CV, while providing accurate estimates of performance and confidence intervals. Finally, we examine the role of repeating the procedure with different partitions to folds ({\em Repeated BBC-CV}) and show that multiple repeats improve the selection of the best configuration (tuning) and lead to better performing models. In addition, for the same number of trained models, Repeated BBC-CV leads to better performing models than NCV while having similar bias in their performance estimates. 

The rest of this paper is structured as follows. In Section~\ref{sec:preliminaries} we present and discuss widely established protocols for tuning and out-of-sample performance estimation. In Section~\ref{sec:related} we discuss additional related work. We introduce our methods BBC-CV and BCED-CV in Sections~\ref{sec:bbc_cv}~and~\ref{sec:bced_cv}, respectively, and empirically evaluate them on synthetic and real settings in Section~\ref{sec:evaluation}. We conclude the paper in Section~\ref{sec:conclusions}.

\section{Preliminaries of Out-of-Sample Estimation}
\label{sec:preliminaries}

In this section, we present the basics of out-of-sample estimation of the performance of a learning method $f$ and introduce the notation employed in the rest of the paper. We assume the learning method is a function that accepts as input a dataset $D = \{\langle x_j, y_j\rangle \}_{j=1}^N$ of pairs of training vectors $x$ and their corresponding labels $y$ and returns another function $M(x)$ (a predictive model), so that $f(D) = M$. We can also think of $D$ as a 2D matrix with the rows containing the examples, and the columns corresponding to features (a.k.a. variables, attributes, measured/observed quantities). It is convenient to employ the {\em Matlab} index notation on matrices to denote with $D(:, i)$ the $i$-th column of $D$ and $D(i, :)$ the $i$-th row of $D$; similarly $D(I, i)$ is the vector of values in the $i$-th column from rows with indexes in vector $I$.

We also overload the notation and use $f(x, D)$ to denote the output (predictions) of the model $M$ trained by $f$ on dataset $D$ when given as input one or multiple samples $x$. We also denote the loss (metric of error) between the value $y$ of a label and a corresponding prediction $\hat{y}$ as $l(y, \hat{y})$. For convenience, we can also define the loss between a {\em vector} of labels $y$ and a {\em vector} of predictions $\hat{y}$ as the vector of losses between the corresponding labels and predictions: 
$$[l(y, \hat{y})]_j = l(y_j, \hat{y_j}) $$ 
The loss function can be either the 0-1 loss for classification (i.e, one when the label and prediction are equal and zero otherwise), the squared error $(y-\hat{y})^2$ for regression or any other metric. Some metrics of performance such as the AUC or the Concordance Index for survival analysis problems \cite{harrell1982} cannot be expressed using a loss function defined on single pairs $\langle y, \hat{y}\rangle$. These metrics can only be computed on a test set containing at least 2 predictions and thus, $l(y, \hat{y}$) is defined only when $y$ and $\hat{y}$ are vectors for such metrics.

\begin{algorithm}[t]
\caption{{\bf CV}($f, D = \{F_1, \ldots, F_K\}$): Basic K-Fold Cross Validation}
\label{alg:CV}
\textbf{Input}: Learning method $f$, Data matrix $D = \{\langle x_j, y_j\rangle \}_{j=1}^N$ partitioned into about equally-sized folds $F_i$\\
\textbf{Output}: Model $M$, Performance estimation $L_{CV}$, out-of-sample predictions $\Pi$ on all folds
\begin{algorithmic}[1]
    \STATE Define $D_{\setminus i} \leftarrow D \setminus F_i$
    \STATE // Obtain the indexes of each fold
    \STATE $I_i \leftarrow \mathit{indexes}(F_i)$
    \STATE // Final Model trained by $f$ on all available data
    \STATE $M \leftarrow f(D)$
    \STATE // Performance estimation: learn from $D_{\setminus i}$, estimate on $F_i$
    \STATE $L_{CV} \leftarrow \frac{1}{K} \sum_{i=1}^K l(y(I_i), f(F_i, D_{\setminus i}))$
    \STATE // Out-of-sample predictions are used by bias-correction methods
    \STATE Collect out-of-sample predictions $\Pi = [f(F_1, D_{\setminus 1}) ; \cdots ; f(F_K, D_{\setminus K})]$
    \STATE {\bf Return } $\langle M, L_{CV}, \Pi \rangle$
\end{algorithmic}
\end{algorithm}

The $K$-fold Cross-Validation (CV) protocol is arguably the most common out-of-sample performance estimation protocol for relatively small sample sizes. It is shown in Algorithm \ref{alg:CV}. The protocol accepts a learning method $f$, a dataset $D$ already partitioned into $K$ folds $F$. {\em The model to return is computed by applying the learning method $f$ on all available data}. To estimate the performance of this model CV employs each fold $F_i$ in turn as an estimation set and trains a model $M_i$ on the remaining data (in the algorithm denoted as $D_{\setminus i}$) using $f$, i.e., $M_i = f(D_{\setminus i})$. It then computes the loss of $M_i$ on the hold-out fold $F_i$. The final performance estimate is the average loss over all folds. The pseudo-code in Algorithm \ref{alg:CV} as presented, also collects and returns all out-of-sample predictions in a vector $\Pi$. This facilitates the presentation of some bias-correction methods below, who depend on them. In case no bias-correction is applied afterwards, $\Pi$ can be omitted from the output arguments.

As simple and common as CV is, there are still several misconceptions about its use. First, the protocol returns $f(D)$ learned from the full dataset $D$, but the losses computed are on different models, namely models trained with subsets of the data. So, CV does not estimate the loss of the specific returned model. We argue that {\em Cross Validation estimates the average loss of the models produced by $f$ when trained on datasets of size $|D_{\setminus i}|$ on the distribution of $D$}. The key conceptual point is that {\em it is not the returned model who is being cross-validated, but the learning method $f$}. Non-expert analysts (and students in particular) often wonder which out of the $K$ models produced by cross-validation excluding each fold in turn should be returned. The answer is none; the model to use operationally is the one learned by $f$ on all of $D$.

Notice that the size of the training sets in CV is $|D_{\setminus i}| = (K-1)/K \cdot |D| < |D|$ (e.g., for 5-fold, we are using only 80\% of the total sample size for training in each fold). It follows that {\em CV estimates are conservatively biased}: the final model $f$ is trained on $|D|$ samples while the estimates are produced by models trained on $|D_{\setminus i}|$ samples. A typical major assumption is that {\em $f$'s models improve on any given task with increased sample size}. This is a reasonable assumption to make, although not always necessarily true. If it does hold, then we expect that the returned loss estimate $L$ of CV is {\em conservative}, i.e,. on average higher than the average loss of the returned model. Exactly how conservative it will be depends on where the classifier is operating on its learning curve for this specific task, which is unknown a priori. It also depends on the number of folds $K$: the larger the $K$, the more $(K-1)/K$ approaches 100\% and the bias disappears. When sample sizes are small or distributions are imbalanced (i.e., some classes are quite rare in the data), we expect most classifiers to quickly benefit from increased sample size, and thus, ̂for CV to be more conservative. 

Based on the above, one expects that leave-one-out CV (where each fold's size is 1 sample) should be the least biased. However, leave-one-out CV can collapse in the sense that it can provide extremely misleading estimates in degenerate situations (see \cite{witten2016data}, p. 151, and \cite{kohavi1995} for an extreme failure of leave-one-out CV and of the $0.632$ bootstrap rule). We believe that the problem of leave-one-out CV stems from the fact that the folds may follow a totally different distribution than the distribution of the class in the original dataset: when only one example is left out, the distribution of one class in the fold is 100\% and 0\% for all the others. We instead advice to use $K$ only as large as possible to still allow the distribution of classes in each fold to be approximately similar as in the original dataset, and impose this restriction when partitioning to folds. The latter restriction leads to what is called {\em stratified CV} and there is evidence that it leads to improved performance estimations \cite{amin2014}.

\subsection{Cross-Validation with Tuning (CVT)}

A typical data analysis involves several algorithms to be combined, e.g., for transforming the data, imputing the missing values, variable selection or dimensionality reduction, and modeling. There are hundreds of choices of algorithms in the literature for each type of algorithms. In addition, each algorithm typically takes several hyper-parameter values that should be tuned by the user. We assume that the learning method $f(D)$ is augmented to $f(D, \theta)$ to take as input a vector $\theta$ that determines which combination of algorithms to run and with what values of hyper-parameters. We call $\theta$ a {\em configuration} and refer to the process of selecting the best $\theta$ as {\em tuning} of the learning pipeline.

The simplest tuning procedure is to cross-validate $f$ with a different configuration $\theta$ each time within a predetermined set of configurations $\Theta$, choose the best performing configuration $\theta^\star$ and then train a final model on all data with $\theta^\star$. The procedure is shown in Algorithm \ref{alg:CVT}. In the pseudo-code, we compute $f_i$ as the closure of $f$\footnote{The term closure is used in the programmatic sense to denote a function produced by another function by binding some free parameters to specific values; see also http://gafter.blogspot.gr/2007/01/definition-of-closures.html.} when the configuration input parameter is grounded to the specific values in $\theta_i$. For example, if configuration $\theta_i$ is a combination of a feature selection algorithm $g$ and modeling algorithm $h$ with their respective hyper-parameter values $a$ and $b$, taking the closure and grounding the hyper-parameters produces a function $f_i = h(g(\cdot,a),b)$, i.e., a function $f_i$ that first applies the specified feature selection $g$ using hyper-parameters $a$ and uses the result to train a model using $h$ with hyper-parameters $b$.
The use of the closures leads to a compact pseudo-code implementation of the method. 

\begin{algorithm}[t]
\caption{{\bf CVT}($f, D = \{F_1, \ldots, F_K\}, \Theta$): Cross Validation With Tuning}
\label{alg:CVT}
\textbf{Input}: Learning method $f$, Data matrix $D = \{\langle x_j, y_j\rangle \}_{j=1}^N$ partitioned into about equally-sized folds $F_i$, set of configurations $\Theta$\\
\textbf{Output}: Model $M$, Performance estimation $L_{CVT}$, out-of-sample predictions $\Pi$ on all folds for all configurations
\begin{algorithmic}[1]
	\FOR{$i=1$ to $C=|\Theta|$}
    \STATE // Create a closure of $f$ (a new function) by grounding the configuration $\theta_i$
    \STATE $f_i \leftarrow$ Closure($f(\cdot, \theta_i)$)
    \STATE $\langle M_i, L_i, \Pi_i \rangle \leftarrow \mathbf{CV}(f_i, D)$
    \ENDFOR
	\STATE $i^\star \leftarrow \arg\min_i L_i$    
    \STATE // Final Model trained by $f$ on all available data using the best configuration
    \STATE $M \leftarrow f(D, \theta_{i^\star})$
    \STATE // Performance estimation; may be optimistic and should not be reported in general
    \STATE $L_{CVT} \leftarrow L_{i^\star}$
    \STATE // Out-of-sample predictions are used by bias-correction methods
    \STATE Collect all out-of-sample predictions of all configurations in one matrix $\Pi \leftarrow [\Pi_1 \cdots \Pi_C ]$
    \STATE {\bf Return } $\langle M, L_{CVT}, \Pi \rangle$
\end{algorithmic}
\end{algorithm}

We now put together two observations already noted above: the performance estimate $L_{CVT}$ of the winning configuration tends to be conservative because it is computed by models trained on only a subset of the data; at the same time, it tends to be optimistic because it is selected as the best among many tries. Which of the two trends will dominate depends on the situation and is a priori unknown. For large $K$ and a large number of configurations tried, the training sets are almost as large as the whole dataset and the optimistic trend dominates. In general, for small sample sizes and a large number of configurations tried {\em $L_{CVT}$ is optimistic and should not be reported as the performance estimate of the final model}. 

\subsection{The Nested Cross-Validation (NCV) Protocol}

\begin{algorithm}[t]
\caption{{\bf NCV}($f, D = \{F_1, \ldots, F_K\}, \Theta$): Nested Cross Validation}
\label{alg:NCV}
\textbf{Input}: Learning method $f$, Data matrix $D = \{\langle x_j, y_j\rangle \}_{j=1}^N$ partitioned into about equally-sized folds $F_i$, set of configurations $\Theta$\\
\textbf{Output}: Model $M$, Performance estimation $L_{NCV}$, out-of-sample predictions $\Pi$ on all folds for all configurations
\begin{algorithmic}[1]
	\STATE // Create closure by grounding the $f$ and the $\Theta$ input parameters of {\bf CVT}
    \STATE $f' \leftarrow \mathbf{CVT}(f, \cdot, \Theta)$
	\STATE // Notice: final Model is trained by $f'$ on all available data; final estimate is provided by basic CV (no tuning) since $f'$ returns a single model each time
    \STATE $\langle M, L_{NCV}, \Pi \rangle \leftarrow \mathbf{CV}(f', D)$
    \STATE {\bf Return } $\langle M, L_{NCV} \rangle$
\end{algorithmic}
\end{algorithm}

Given the potential optimistic bias of CV when tuning takes place, other protocols have been developed, such as the Nested Cross Validation (NCV). We could not trace who introduced or coined up first the name Nested Cross-Validation but the authors and colleagues have independently discovered it and using it since 2005 \cite{statnikov2005}; around the same time Varma and Simon in \cite{varma2006}, report a bias in error estimation when using K-Fold Cross-Validation, and suggest the use of the Nested K-Fold Cross-Validation (NCV) protocol as an almost unbiased estimate of the true performance. A similar method in a bioinformatics analysis was used in 2005 \cite{statnikov2005}. One early comment hinting of the method is in \cite{salzberg1997comparing}, while Witten and Frank (see \cite{witten2005}, page 286) briefly discuss the need of treating any parameter tuning step as part of the training process when assessing performance. It is interesting to note that the earlier works on NCV appeared first in bioinformatics where the sample size of datasets is often quite low and the effects of the bias more dramatic.

The idea of the NCV is evolved as follows. Since the tuning sets have been used repeatedly for selecting the best configuration, one needs a second hold-out set exclusively for estimation of one, single, final model. However, one could repeat the process with several held-out folds and average the estimates. In other words, each fold is held-out for estimation purposes each time and a CVT takes place for the remaining folds in selecting the best configuration and training on all remaining data with this best configuration to return a single model. Thus, in NCV each fold serves once for estimation and multiple times as a tuning set. Under this perspective, NCV is a generalization of a double-hold-out protocol partitioning the data to train-tuning-estimation.

Another way to view NCV is to {\em consider tuning as part of the learning process}. The result is a new learning function $f'$ that returns a single model, even though internally it is using CV to select the best configuration to apply to all input data. NCV simply cross-validates $f'$. What is this new function $f'$ that uses CV for tuning and returns a single model? It is actually CVT for the given learning method $f$ and configuration set $\Theta$. Naturally, any method that performs hyper-parameter optimization and returns a single model can be used instead of CVT as $f'$. The pseudo-code in Algorithm \ref{alg:NCV} clearly depicts this fact and implements NCV in essentially two lines of code using the mechanism of closures.

Counting the number of models created by NCV, let us denote with $C = |\Theta|$ the number of configurations to try. To produce the final model, NCV will run CVT on all data. This will create $K\times C$ models for tuning and once the best configuration is picked, one more model will be produced leading to $K\times C + 1$ models for final model production. To produce the estimate, the whole process is cross-validated each time excluding one fold, thus leaving $K-1$ folds for the inner cross-validation loop (the loop inside $f'$). Overall, this leads to $K \times ((K-1)\times C + 1)$ models trained for estimation. The total count is exactly $K^2\times C + K + 1$ models, which is of course computationally expensive as it depends quadratically on the number of folds $K$.

\subsection{The Tibshirani and Tibshirani Protocol}

\begin{algorithm}[t]
\caption{{\bf TT}($f, D = \{F_1, \ldots, F_K\}, \Theta$): Cross Validation with Tuning, Bias removal using the TT method}
\label{alg:TT}
\textbf{Input}: Learning method $f$, Data matrix $D = \{\langle x_j, y_j\rangle \}_{j=1}^N$ partitioned into about equally-sized folds $F_i$, set of configurations $\Theta$\\
\textbf{Output}: Model $M$, Performance estimation $L_{TT}$
\begin{algorithmic}[1]
	\STATE // Notice: the final Model is the same as in CVT	
    \STATE $\langle M, L_{CVT}, \Pi \rangle \leftarrow \mathbf{CVT}(f, D, \Theta)$
    \FOR{$k=1$ to $K$}
    %\STATE $I_k \leftarrow \mathit{indexes}(F_k)$		// Obtain indexes of rows of fold $k$
    \STATE // Compute bias estimate for fold $k$
   	\STATE $\mathit{TTBias_k} \gets l(y(I_k), \Pi(I_k,j)) - \underset{i}{min}\, l(y(I_k), \Pi(I_k,i))$
    \ENDFOR
    \STATE $\mathit{TTBias} \gets \frac{1}{K} \sum_{k=1}^K \mathit{TTBias}_k$
    \STATE $ L_{TT} \leftarrow L_{CVT} + \mathit{TTBias}$
    \STATE {\bf Return } $\langle M, L_{TT} \rangle$
\end{algorithmic}
\end{algorithm}

To reduce the computational overhead of NCV, Tibshirani and Tibshirani \cite{tibshirani2009} introduced a new method for estimating and correcting for the bias of CVT without training additional models. We refer to this method as the TT and it is the first work of its kind, inspiring this work.

The main idea of the TT method is to consider, in a sense, each fold a different dataset and serving as an independent example to estimate how much the process of selecting the best configuration out of many incurs optimism. It compares the loss of the final, selected configuration with the one selected in a given fold as an estimate of the bias of the selection process. Let $I_k$ denote the indexes of the samples (rows) of the $k$-th fold $F_k$. Furthermore, let $j$ denote the index of the best performing configuration (column of $\Pi$), as computed by CVT.
The bias $\mathit{TTBias}$ estimated by the TT method is computed as:
$$
\mathit{TTBias} = \frac{1}{K} \sum_{k=1}^{K} (l(y(I_k), \Pi(I_k,j)) - \min_i l(y(I_k), \Pi(I_k,i)))
$$
Note that, the average of the first terms $l(y(I_k), \Pi(I_k,j))$ in the sum is the average loss of the best configuration computed by CVT, $L_{CVT}$. Thus, $\mathit{TTBias}$ can be rewritten as:
$$
\mathit{TTBias} = L_{CVT} - \frac{1}{K} \sum_{k=1}^{K} \min_i l(y(I_k), \Pi(I_k,i))
$$
The final performance estimate is:
$$
L_{TT} = L_{CVT} + \mathit{TTBias}
$$
The pseudo-code is presented in Algorithm \ref{alg:TT} where it is clear that the TT does not train new models, employs the out-of-sample predictions of all models and corresponding configurations, and returns the same final model as both the CVT and the NCV. It is also clear that when the same configuration is selected on each fold as the final configuration, the bias estimate is zero.

A major disadvantage of the TT is also apparent. Observe that the bias estimate of TT obeys $0 \leq \mathit{TTBias} \leq L_{CVT}$. Thus, the final estimate $L_{TT}$ is always between $L_{CVT}$ and $2L_{CVT}$. This can trivially lead to cases where TT over-corrects the loss or does not perform any correction at all. As an example of the former, consider the extreme case of classification, 0-1 loss and Leave-One-Out CV where each fold contains a single instance. Then it is likely, especially if many configurations have been tried, that there always is a configuration that correctly predicts the held-out sample in each fold. Thus, in this scenario the bias estimate will be exactly equal to the loss of the selected configuration and so $L_{TT} = 2L_{CVT}$. If for example in a multi-class classification problem, the selected configuration has an estimated 0-1 loss of 70\%, the TT method will adjust it to return 140\% loss estimate! If on the other hand the 0-1 loss is close to zero, almost no correction will be performed by TT. Such problems are very likely to be observed with few samples and if many configurations are tried. For reliable estimation of the bias, the TT requires relatively large folds, but it is exactly the analyses with overall small sample size that need the bias estimation the most. For the same reason, it is less reliable for performance metrics such as the AUC or the concordance index (in survival analysis) that require several predictions to be computed; thus, estimating these metrics in small folds is totally unreliable.

\section{Related Work}
\label{sec:related}

There are two additional major works that deal with performance estimation when tuning (model selection) is included. 
Bernau et al. \cite{bernau2013} introduced two variants of a bias correction method as a smooth analytical alternative to NCV, the WMC and WMCS. The method is based on repeated subsampling of the original dataset and training of multiple models. It then computes the error estimate as a weighted mean of the error rates of every configuration over all the subsamples. The two variants differ in the way that the weights are calculated. Compared to NCV, the authors claim that WMC/WMCS is competitive and more stable, for the same number of trained models as the CVT. 
However, subsequent independent work by \cite{ding2014} report problems with the method, and specifically that it provides fluctuating estimates and it may over-correct the bias in some cases. It is also complicated to understand and implement.

Ding et al. in \cite{ding2014} proposed a resampling-based inverse power law (IPL) method for bias correction and compared its performance to those of TT, NCV, and WMC/WMCS on both simulated and real datasets. The error rate of each classifier is estimated by fitting a learning curve which is constructed from repeatedly resampling the original dataset for different sample sizes and fitting an inverse power law function. The IPL method outperforms the other methods in terms of performance estimation but, as the authors point out, it exhibits significant limitations. Firstly, it is based on the assumption that the learning curve for each classifier can be fitted well by inverse power law. Additionally, if the sample size of the original dataset is small, the method will provide unstable estimates. Lastly, the IPL method has higher computational cost compared to TT and the WMC/WMCS methods.

\section{The Bootstrap Bias Corrected Cross-Validation (BBC-CV)}
\label{sec:bbc_cv}

The bootstrap \cite{efron1993} has been developed and applied extensively to estimate in a non-parametric way the (unknown) distribution of a statistic $b_o$ computed for a population (dataset). The main idea of the bootstrap is to sample with replacement from the given dataset multiple times (e.g., 500), each time computing the statistic $b_i, i=1, \ldots, B$ on the resampled dataset. The empirical distribution of $b_i$, under certain broad conditions approaches the unknown distribution of $b_o$. Numerous variants have appeared for different statistical tasks and problems (see \cite{davison1997}).

In machine learning, for estimation purposes the idea of bootstrapping datasets has been proposed as an alternative to the CV. Specifically, to produce a performance estimate for a method $f$ multiple training sets are produced by bootstrap (uniform re-sampling with replacement of rows of the dataset), a model is trained and its performance is estimated on the out-of-sample examples. On average, random re-sampling with replacement results in 63.2\% of the original samples included in each bootstrap dataset and the rest serving as out-of-sample test sets. The protocol has been compared to the CV in \cite{kohavi1995} concluding that the CV is preferable.% to the bootstrap as described. Perhaps, due to this result the CV was never replaced by the bootstrap in practice.

The setting we explore in this paper is different than what described above since we examine the case where one is also tuning. A direct application of the bootstrap idea in such settings would be to substitute CVT (instead of CV) with a bootstrap version where not one but all configurations are tried on numerous bootstrap datasets, the best is selected, and its performance is estimated as the average loss on the out-of-sample predictions. Obviously, this protocol would require the training of $B\times C$ models, where $B$ is the number of bootstraps, an unacceptably high computational overhead for $B$ in the typical range of a few hundreds to thousands.

Before proceeding with the proposed method, let us define a new important function {\bf css}$(\Pi, y)$ standing for {\em configuration selection strategy}, where $\Pi$ is a matrix of out-of-sample predictions and $y$ is a vector of the corresponding true labels.
Recall that $\Pi$ contains $N$ rows and $C$ columns, where $N$ is the sample size and $C$ is the number of configurations so that $[\Pi]_{ij}$ denotes the out-of-sample prediction of on the $i$-th sample of the $j$-th configuration. The function {\bf css} returns the index of the best-performing configuration according to some criterion. The simplest criterion, also employed in this paper, is to select the configuration with the minimum average loss:
$$
\mathbf{css}(\Pi, y) = \arg\min_i l(y, \Pi(:, i))
$$
where we again employ the Matlab index notation $\Pi(:, i)$ to denote the vector in column $i$ of matrix $\Pi$, i.e,. all pooled out-of-sample predictions of configuration $i$. However, by explicitly writing the selection as a new function, one can easily implement other selection criteria that consider, not only the out-of-sample loss, but also the complexity of the models produced by each configuration. 

We propose the Bootstrap Bias Corrected CV method (BBC-CV), for efficient and accurate performance estimation.
The pseudo-code is shown in Algorithm~\ref{alg:BBC}.
BBC-CV uses the out-of-sample predictions $\Pi$ returned by CVT.
It creates $B$ bootstrapped matrices $\Pi^b, b = 1, \dots, B$ and the corresponding vectors of true labels $y^b$ by sampling $N$ rows of $\Pi$ with replacement. Let $\Pi^{\setminus b}, b = 1, \dots, B$ denote the matrices containing the samples in $\Pi$ and not in $\Pi^b$ (denoted as $\Pi \setminus \Pi^b$), and $y^{\setminus b}$ their corresponding vectors of true labels.
%Let $out^b$ denote the indices of the samples not included in $\Pi^b$.
For each bootstrap iteration $b$, BBC-CV: (a) applies the configuration selection strategy css($\Pi^b, y^b$) to select the best-performing configuration $i$, and (b) computes the loss $L_b$ of configuration $i$ as $L_b = l(y^{\setminus b}, \Pi^{\setminus b})$. Finally, the estimated loss $L_{BBC}$ is computed as the average of $L_b$ over all bootstrap iterations.

\begin{algorithm}[t]
\caption{{\bf BBC-CV}($f, D = \{F_1, \ldots, F_K\}, \Theta$): Cross Validation with Tuning, Bias removal using the BBC method}
\label{alg:BBC}
\textbf{Input}: Learning method $f$, Data matrix $D = \{\langle x_j, y_j\rangle \}_{j=1}^N$ partitioned into approximately equally-sized folds $F_i$, set of configurations $\Theta$\\
\textbf{Output}: Model $M$, Performance estimation $L_{BBC}$, 95\% confidence interval $[lb, ub]$
\begin{algorithmic}[1]
	\STATE // Notice: the final Model is the same as in CVT
    \STATE $\langle M, L_{CVT}, \Pi \rangle \leftarrow \mathbf{CVT}(f, D, \Theta)$
    \FOR{$b=1$ to $B$}
    \STATE $\Pi^b \leftarrow $ sample with replacement $N$ rows of $\Pi$
    \STATE $\Pi^{\setminus b} \leftarrow \Pi \setminus \Pi^b$ // get samples in $\Pi$ and not in $\Pi^b$
    \STATE // Apply the configuration selection method on the bootstrapped out-of-sample predictions
    %\STATE $i \leftarrow \mathbf{ccs}(\Pi(\mathit{in}, :), y(\mathit{in}))$
    \STATE $i \leftarrow \mathbf{ccs}(\Pi^b, y^b)$
    \STATE // Estimate the error of the selected configuration on predictions not selected by this bootstrap
    \STATE $L_b \leftarrow l(y^{\setminus b}, \Pi^{\setminus b})$
    \ENDFOR
    \STATE $ L_{BBC} = \frac{1}{B}\sum_{b=1}^B L_b$
    \STATE // Compute 95\% confidence interval; $b_{(k)}$ denotes the $k$-th value of $b$'s in ascending order
    \STATE $[lb, ub] = [b_{(0.025\cdot B)}, b_{(0.975\cdot B)}]$
    \STATE {\bf Return } $\langle M, L_{BBC}, [lb, ub]\rangle$
\end{algorithmic}
\end{algorithm}

BBC-CV differs from the existing methods in two key points. (a) the data that are being bootstrapped are in the matrix $\Pi$ of the pooled out-of-sample predictions computed by CVT (instead of the actual data in $D$), and (b) the method applied on each bootstrap sample is the configuration selection strategy \textbf{css} (not the learning method $f$). Thus, performance estimation can be applied with minimal computational overhead, as {\em no new models need to be trained}.

A few comments on the BBC-CV method now follow. First, notice that if a single configuration is always selected as best, the method will return the bootstrapped mean loss of this configuration instead of the mean loss on the original predictions. The first asymptotically approaches the second as the number of bootstrap iterations increase and they will coincide. A single configuration may always selected for two reasons: either only one configuration was cross-validated or one configuration dominates all others with respect to the selection criterion. In both these cases the BBC-CV estimate will approach the CVT estimate.

Second, BBC-CV simultaneously considers a bootstrap sample from all predictions of all configurations, not only the ones pertaining to a single fold each time. Thus, unlike TT, it is robust even when folds contain only one or just a few samples. For the same reason, it is also robust when the performance metric is the AUC (or a similar metric) and requires multiple predictions to be computed reliably. There is one caveat however, with the use of BCC-CV and the AUC metric: {\em because BBC-CV pools together predictions from different folds, and thus different models (although produced with the same configuration), the predictions in terms of scores have to be comparable (in the same scale) for use with the AUC}. Finally, we note that we presented BBC in the context of $K$-fold CV, but the main idea of bootstrapping the pooled out-of-sample predictions of each configuration can be applied to other protocols. One such protocol is the hold-out where essentially there is only one fold. Similarly, it may happen that an implementation of $K$-fold CV, to save computational time decides to terminate only after a few folds have been employed, e.g., because the confidence intervals of performance are tight enough and there is no need to continue. We call the latter the {\em incomplete CV protocol}. Again, even though predictions are not available for all samples, BBC-CV can be applied to the predictions of any folds that have been employed for tuning.

\subsection{Computing Confidence Intervals with the Bootstrap}
\label{sec:conf_intervals}

The idea of bootstrapping the out-of-sample predictions can not only correct for the bias, but also trivially be applied to provide confidence intervals of the loss. $1-\alpha$ (commonly 95\%) confidence intervals for a statistic $b_0$ are provided by the bootstrap procedure by computing the population of bootstrap estimates of the statistics $b_1, \ldots, b_B$ and considering an interval $[lb, ub]$ that contains $p$ percentage of the population \cite{efron1993}. The parameter $1-\alpha$ is called the confidence level of the interval. The simplest approach to compute such intervals is to consider the ordered statistics $b_{(1)}, \ldots, b_{(B)}$, where $b_{i}$ denotes the $i$-th value of $b$'s in ascending order, and take the interval $[b_{(\alpha/2 \cdot B)}, b_{((1-\alpha/2) \cdot B)}]$, excluding a probability mass of $\alpha/2$ on each side of extreme values. For example, when $\alpha = 0.05$ and $B=1000$ we obtain $[lb, ub] = [b_{(25)}, b_{(975)}]$. Other variants are possible and could be applied, although outside the scope of this paper. For more theoretical details on the bootstrap confidence intervals and different methods for constructing them, as well as a comparison of them, see \cite{efron1993}.

\subsection{BCC-CV with Repeats}

When sample size is small, the variance of the estimation of the performance is large, even if there is no bias. This is confirmed in \cite{amin2014} empirically on several real datasets. A component of the variance of estimation stems from the specific random partitioning to folds. To reduce this component it is advisable to repeat the estimation protocol multiple times with several fold partitions, leading to the Repeated Cross Validation protocol and variants.  

Applying the BBC-CV method with multiple repeats is possible with the following minimal changes in the implementation: We now consider the matrix $\Pi$ of the out-of-sample predictions of the models to be three dimensional with $[\Pi]_{ijk}$ to denote the out-of-sample prediction (i.e, when the example was held-out during training) on the $i$-th example, of the $j$-th configuration, in the $k$-th repeat. Note that {\em predictions for the same instance $x_i$ in different repeats are correlated}: they all tend to be precise for easy-to-predict instances and tend to be wrong for outliers that do not fit the assumptions of the configuration correctly. Thus, predictions on the same instance for different repeats have to all be included in a bootstrap sample or none at all. In other words, as in Algorithm \ref{alg:BBC}, what is resampled with replacement to create the bootstrap data are the indexes of the instances. Other than that, the key idea remains the same as in Algorithm \ref{alg:BBC}.

\section{Bootstrap Corrected with Early Dropping Cross-Validation (BCED-CV)}
\label{sec:bced_cv}

In this section, we present a second use of the idea to bootstrap the pooled out-of-sample predictions of each configuration. Specifically, they are employed as part of a statistical hypothesis test that determines whether a configuration's performance is statistically significantly inferior than the performance of the current best configuration. If this is indeed the case, the dominated configuration can be early dropped from further consideration, in the sense that no additional models on subsequent folds will be trained under this configuration. If a strict significance threshold is employed for the test then the dropped configurations have a low probability of actually ending up as the optimal configuration at the end of the CVT and thus, the prediction performance of the final model will not be affected. The Early Dropping scheme can lead to substantial computational savings as numerous configurations can be dropped after just a few folds before completing the full $K$-fold CV on them. 

Specifically, the null hypothesis of the test is that a given configuration's $\theta$ performance (loss) is equal to the current best configuration $\theta_o$, i.e., $H_\theta: l_N(\theta) = l_N(\theta_o)$, where $l$ is the average loss of the models produced by the given configuration when trained on datasets from the distribution of the problem at hand of size $N$. Since all models are produced by the same dataset size stemming from excluding a single fold, we can actually drop the subscript $N$. These hypotheses are tested for every $\theta$ that is still under consideration at the end of each fold
%$\footnote{When folds are quite small, e.g,. in the case of leave-one-out cross validation, testing could take place every $L>1$ folds instead: including just a single additional prediction in the test will not result in any new methods dropping.}, 
 i.e., as soon as new out-of-sample predictions are accrued for each configuration. 

To perform the test, the current, pooled, out-of-sample predictions of all configurations still under consideration $\Pi$ are employed to identify the best current configuration $\theta_o = \mathbf{css}(\Pi, y)$. Subsequently, $\Pi$'s rows are bootstrapped to create matrices $\Pi^1, \ldots, \Pi^B$ and corresponding label matrices $y^1, \ldots, y^B$. From the population of these bootstrapped matrices the probability $p_\theta$ of a given configuration $\theta$ to exhibit a worse performance than $\theta_o$ is estimated as the percentage of times its loss is higher than that of $\theta$'s, i.e., $\hat{p}_\theta = \frac{1}{B}\#\{l(y^b, \Pi^b(:, \theta)) > l(y^b, \Pi^b(:, \theta_o)), b=1, \ldots, B\}$. If $\hat{p}_\theta > \alpha$ for some significance threshold (e.g., $\alpha = 0.99$), configuration $\theta$ is dropped. 

A few comments on the procedure above. It is a heuristic procedure mainly with focus on computational efficiency, not statistical theoretical properties. Ideally, the null hypothesis to test for each configuration $\theta$ would be the hypothesis that $\theta$ will be selected as the best configuration at the end of the CVT procedure, given a finite number of folds remain to be considered. If this null hypothesis is rejected for a given $\theta$, $\theta$ should be dropped. Each of these hypotheses for a given $\theta$ has to be tested in the context of all other configurations that participate in the CVT procedure. In contrast, the heuristic procedure we provide essentially tests each hypothesis $H_\theta$ in isolation. For example, it could be the case during bootstrapping, configuration $\theta$ exhibits a significant probability of a better loss than $\theta_o$ (not dropped by our procedure), but it could be that in all of these cases, it is always dominated by some other configuration $\theta'$. Thus, the actual probability of being selected as best in the end maybe smaller than the percentage of times it appears better than $\theta_o$. 

In addition, our procedure does not consider the uncertainty (variance) of the selection of the current best method $\theta_o$. Perhaps, a double bootstrap procedure would be more appropriate in this case \cite{nankervis2005} but any such improvements would have to also minimize the computational overhead to be worthwhile in practice. 

\subsection{Related work}
\label{subsec:related}
The idea of accelerating the learning process by specifically eliminating under-performing configurations from a finite set, early within the cross-validation procedure, was introduced as early as 1994 by Maron and Moore with \textit{Hoeffding Races} \cite{maron1994hoeffding}. At each iteration of leave-one-out CV (i.e. after the evaluation of a new test point) the algorithm employs the Hoeffding inequality for the construction of confidence intervals around the current error rate estimate of each configuration. Configurations whose intervals do not overlap with those of the best-performing one, are eliminated (dropped) from further consideration. The procedure is repeated %until one configuration has been left, the data points have all served as test points, or 
until the confidence intervals have shrunk enough so that a definite overall best configuration can be identified. %The intervals get smaller as more data points are tested,
However, many test point evaluations may be required before a configuration can clearly be declared the winner.

Following a similar approach, Zheng and Bilenko in 2013 \cite{zheng2013lazy} applied the concept of early elimination of suboptimal configurations to K-fold CV. They improve on the method by Maron and Moore by incorporating paired hypothesis tests for the comparison of configurations for both discrete and continuous hyper-parameter spaces. At each iteration of CV, all current configurations are tested pairwise and those which are inferior are dropped. Then, power analysis is used to determine the number of new fold evaluations for each remaining configuration given an acceptable false negative rate.

Krueger \textit{et al.} \cite{krueger2015} in 2015 introduced the so-called \textit{Fast Cross-Validation via Sequential
Testing} (CVST) which uses nonparametric testing together with sequential analysis in order to choose the best performing configuration on the basis of linearly increasing subsets of data. At each step, the Friedman \cite{friedman1937} or the Cochran's~Q test \cite{cochran1950} (for regression and classification tasks respectively) are employed in order to detect statistically significant differences between configurations' performances. Then, the seemingly under-performing configurations are further tested through sequential analysis to determine which of them will be discharged. Finally, an early stopping criterion is employed to further speed~up the CV process. The winner configuration is the one that has the best average ranking, based on performance, in the last few iterations specified in advance. The disadvantage of CVST is that it initially operates on smaller subsets, thus risking the early elimination of good models when the original dataset is already small.   

In none of the methods above the bias of the performance estimate of a partially completed CV is examined. Our approach, BCED-CV, utilizes the bootstrap correction protocol (BBC-CV) and provides an almost unbiased estimate of performance of the returned model. In comparison to the statistical tests used in \cite{zheng2013lazy} and \cite{krueger2015}, the bootstrap is a general test, applicable to any type of learning task and measure of performance, and is suitable even for relatively small sample sizes. Finally, BCED-CV requires that only the value of the significance threshold $\alpha$ is pre-specified while the methods in \cite{zheng2013lazy} and \cite{krueger2015} have a number of hyper-parameters to be specified in advance.  

\section{Empirical Evaluation}
\label{sec:evaluation}

We empirically evaluate the efficiency and investigate the properties of BBC-CV and BCED-CV, on both controlled settings and real problems. In particular, we focus on the bias of the performance estimates of the protocols, and on computational time. We compare the results to those of three standard approaches: CVT, TT and NCV. We also examine the tuning (configuration selection) properties of BBC-CV, BCED-CV and BBC-CV with repeats, as well as the confidence intervals that these methods construct.  

%assess, examine
\subsection{Simulation studies}

Extensive simulation studies were conducted in order to validate BBC-CV and BCED-CV, and assess their performance. We focus on the binary classification task and use classification accuracy as the measure of performance. %The variable factors of the simulations were
We examine multiple settings for varying sample size $N \in \{20, 40, 60, 80, 100, 500, 1000\}$, number of candidate configurations $C \in \{50, 100, 200, 300, 500, 1000, 2000\}$, and true performances $P$ of the candidate configurations drawn from different Beta distributions $ Be(a,b)$ with $(a, b) \in \{(9, 6), (14, 6), (24, 6), (54, 6)\}$, corresponding to a mean value of $\mu \in \{0.6, 0.7, 0.8, 0.9\}$ and variance of $0.015, 0.01, 0.0052, 0.0015$. These choices result in a total of 196 different experimental settings. We chose distributions with small variances since these are the most challenging cases where the models have more similar performances. 

For each setting, we generate a matrix of out-of-sample predictions $\Pi$. First, a true performance value $P_j, j = 1,\ldots,C$, sampled from the same beta distribution, is assigned to each configuration $c_j$. Then, the sample predictions for each $c_j$ are produced as $\Pi_{ij} = \mathds{1}(r_i < P_j), i = 1,\ldots,N$, where $r_i$ are random numbers sampled uniformly from $(0,1)$, and $\mathds{1}(condition)$ denotes the unit (indicator) function. 

Then, the BBC-CV, BCED-CV, CVT, TT, and NCV protocols for tuning and performance assessment of the returned model are applied. We set the number of bootstraps $B = 1000$ for the BBC-CV method, and for the BCED-CV we set $B = 1000$ and the dropping threshold to $a = 0.99$. We applied the same split of the data into $K = 10$ folds for all the protocols. Consequently, all of them, with the possible exception of the BCED-CV, select and return the same predictive model with different estimations of its performance. The internal cross-validation loop of the NCV uses $K = 9$ folds. 

The whole procedure was repeated 500 times for each setting, leading to a total of 98,000 generated matrices of predictions, on which the protocols were applied. The results presented are the averages over the 500 repetitions.  

\begin{figure}[!t]
\centering
\begin{tabular}{cc}
\includegraphics[scale=0.45,trim=20 0 0 0]{sim_CVT_bias_60.eps} &
\includegraphics[scale=0.45,trim=200 0 0 0]{sim_TT_bias_60.eps}  \\ 
\includegraphics[scale=0.45,trim=20 0 0 0]{sim_NCV_bias_60.eps}  &
\includegraphics[scale=0.45,trim=200 0 0 0]{sim_BBC_bias_60.eps}   \\
\includegraphics[scale=0.45,trim=20 0 0 0]{sim_BCED_bias_60.eps}  &
\includegraphics[scale=0.6,trim=0 80 0 0]{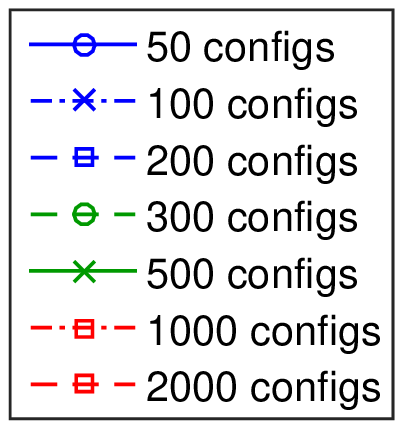}
\end{tabular}
\caption{Average estimated bias (over 500 repeats) of the CVT, TT, NCV, BBC-CV and BCED-CV estimates of performance for an average true classification accuracy of 60\%. CVT over-estimates performance in all settings. TT's behaviour varies for sample size $N < 500$ and is conservative for $N \geq 500$. NCV provides almost unbiased estimates of performance, while BBC-CV is more conservative with a difference in the bias of $0.013$ points of accuracy on average. BCED-CV is on par with NCV.\label{fig:sim_average_bias}}
\end{figure}

%width=0.5\textwidth

\subsubsection{Bias Estimation}

The bias of the estimation is computed as $\widehat{Bias} = \hat{P} - P$, where $\hat{P}$ and $P$ denote the estimated and the true performance of the selected configuration, respectively. A positive bias indicates a lower true performance than the one estimated by the corresponding performance estimation protocol and implies that the protocol is optimistic (i.e. overestimates the performance), whereas a negative bias indicates that the estimated performance is conservative. Ideally, the estimated bias should be 0, although a conservative estimate is also acceptable in practice.

Figure~\ref{fig:sim_average_bias} shows the average estimated bias for models with average true classification accuracy $\mu = 0.6$, over 500 repetitions, of the protocols under comparison. Each panel corresponds to a different protocol (specified in the title) and shows the bias of its performance estimate relatively to the sample size (horizontal axis) and the number of configurations tested (different plotted lines). We omit results for the rest of the tested values of $\mu$ as they are similar.

The CVT estimate of performance is optimistically biased in all settings with the bias being as high as $0.17$ points of classification accuracy. We notice that the smaller the sample size, the more CVT overestimates the performance of the final model. However, as sample size increases, the bias of CVT tends to 0. Finally, we note that the bias of the estimate also grows as the number of models under comparison becomes greater, although the effect is relatively small in this experiment.
The behaviour of TT greatly varies for small sample sizes ($\leq 100$), and is highly sensitive to the number of configurations. On average, the protocol is optimistic (not correcting for the bias of the CVT estimate) for sample size $N \in \{20, 40\}$, and over-corrects, for $N \in \{60, 80, 100\}$. For larger sample size ($\geq 500$), TT is systematically conservative, over-correcting the bias of CVT.
NCV provides an almost unbiased estimation of performance, across all sample sizes. However, recall that it is computationally expensive since the number of models that need to be trained depends quadratically on the number of folds $K$.

BBC-CV provides conservative estimates, having low bias which quickly tends to zero as sample size increases. Compared to TT, it is better fitting for small sample sizes and produces more accurate estimates overall. In comparison to NCV, BBC-CV is somewhat more conservative with a difference in the bias of $0.013$ points of accuracy on average, and $0.034$ in the worst case (for $N = 20$). However, we believe that the much lower computational cost (one order of magnitude) of BBC-CV compensates for its conservatism. BCED-CV displays similar behaviour to BBC-CV, having lower bias which approaches zero faster. It is on par with NCV, having $0.005$ points of accuracy more bias on average, and $0.018$ in the worst case. As we show later on, BCED-CV is up to one order of magnitude faster than CVT, and consequently two orders of magnitude faster than NCV.

In summary, the proposed BBC-CV and BCED-CV methods produce almost unbiased performance estimates, and perform only slightly worse in small sample settings than the computationally expensive NCV.
As expected, CVT is overly optimistic, and thus should not be used for performance estimation purposes.
Finally, the use of TT is discouraged, as (a) its performance estimate varies a lot for different sample sizes and numbers of configurations, and (b) it overestimates performance for small sample sizes, which are the cases where bias correction is needed the most.

\subsection{Real Datasets}

After examining the behaviour of BBC-CV and BCED-CV on controlled settings, we investigated their performance on real datasets. Again we focus on the binary classification task but now we use the AUC as the metric of performance. All of the datasets utilized for the experiments come from popular data science challenges (NIPS 2003 \cite{Guyon2004}, WCCI 2006 \cite{Guyon2006}, ChaLearn AutoML \cite{Guyon2015}). Table~\ref{table:datasets} summarizes their characteristics. The domains of application of the ChaLearn AutoML challenge's datasets  are not known, however the organizers claim that they are diverse and were chosen to span different scientific and industrial fields. \textit{gisette} \cite{Guyon2004} and \textit{gina} \cite{Guyon2006} are handwritten digit recognition problems, \textit{dexter} \cite{Guyon2004} is a text classification problem, and \textit{madelon} \cite{Guyon2004} is an artificially constructed dataset characterized by having no single feature that is informative by itself.

The experimental set-up is similar to the one used by Tsamardinos \textit{et al.} \cite{amin2014}. Each original dataset $D$ was split into two stratified subsets; $D_{pool}$ which consisted of 30\% of the total samples in $D$, and $D_{holdout}$ which consisted of the remaining 70\% of the samples. For each original dataset with the exception of \textit{dexter}, $D_{pool}$ was used to sample (without replacement) 20 sub-datasets for each sample size $N \in \{20, 40, 60, 80, 100, 500\}$. For the \textit{dexter} dataset we sampled 20 sub-datasets for each $N \in \{20, 40, 60, 80, 100\}$. We created a total of $8 \times 20 \times 6 + 20 \times 5 = 1060$ sub-datasets. $D_{holdout}$ was used to estimate the true performance of the final, selected model of each of the protocols tested.

\begin{table*}[!t]
\centering
\caption{Datasets' characteristics. $pr/nr$ denotes the ratio of positive to negative examples in a dataset. $|D_{pool}|$ refers to the portion of the datasets (30\%) from which the sub-datasets were sampled and $|D_{holdout}|$ to the portion (70\%) from which the true performance of a model is estimated.\label{table:datasets}}
\begin{tabular}{@{}lcccccccc@{}}
\toprule
%\multicolumn{1}{l}{Name}
    Name & \#Samples & \#Variables & $pr/nr$ & $|D_{pool}|$ & $|D_{holdout}|$ & Source\\
\midrule
    christine & 5418 & 1636 & 1 & 1625 & 3793 & \cite{Guyon2015} \\
    %\specialcell{ChaLearn AutoML\\Challenge \cite{Guyon2015}} \\
    jasmine & 2984 & 144 & 1 & 895 & 2089 & \cite{Guyon2015} \\
    philippine & 5832 & 308 & 1 & 1749 & 4082 & \cite{Guyon2015} \\
    madeline & 3140 & 259 & 1.01 & 942 & 2198 & \cite{Guyon2015} \\
    sylvine & 5124 & 20 & 1 & 1537 & 3587 & \cite{Guyon2015} \\
    gisette & 7000 & 5000 & 1 & 2100 & 4900 & \cite{Guyon2004} \\
	madelon & 2600 & 500 & 1 & 781 & 1819 & \cite{Guyon2004} \\
	dexter & 600 & 20000 & 1 & 180 & 420 & \cite{Guyon2004} \\
	gina & 3468 & 970 & 1.03 & 1041 & 2427 & \cite{Guyon2006} \\
	%farm ads & 4143 & 54877 & 0.87 & 1243 & 2900 & \cite{Lichman2013} \\
\bottomrule
\end{tabular}
\end{table*}

\begin{figure}[!ht]
\centering
\begin{tabular}{cc}
\includegraphics[scale=0.45,trim=20 0 0 0]{real_CVT_mean_bias_pooling_20.eps} &
\includegraphics[scale=0.45,trim=180 0 0 0]{real_TT_mean_bias_20.eps}  \\ 
\includegraphics[scale=0.45,trim=20 0 0 0]{real_NCV_mean_bias_20.eps}  &
\includegraphics[scale=0.45,trim=180 0 0 0]{real_BBC_mean_bias_20.eps}   \\
\includegraphics[scale=0.45,trim=20 0 0 0]{real_BCED_mean_bias_20.eps}  &
\includegraphics[scale=0.6,trim=0 100 10 0, clip = true]{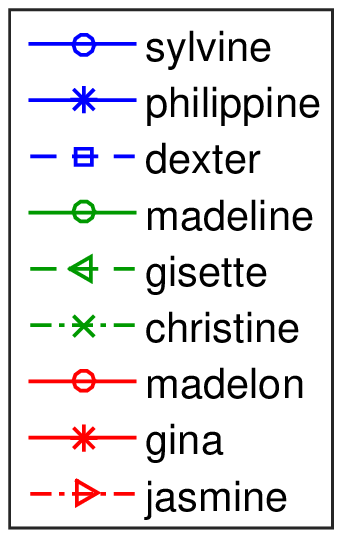}
\end{tabular}
\caption{Average estimated bias (over 20 sub-datasets for each original dataset) of the CVT, TT, NCV, BBC-CV and BCED-CV estimates of performance. CVT is optimistically biased for sample size $N \leq 100$. TT's bias varies with sample size and dataset, and it is mainly over-conservative for $N \geq 80$. NCV and BBC-CV, both have low bias though results vary with dataset. BCED-CV has, on average, greater bias than BBC-CV for $N \leq 100$ and identical for $N = 500$. \label{fig:real_average_bias}}
\end{figure}

The set $\Theta$ (i.e. the search grid) explored consists of 610 configurations. These resulted from various combinations of preprocessing, feature selection, and learning methods and different values for their hyper-parameters. The preprocessing methods included imputation, binarization (of categorical variables) and standardization (of continuous variables) and were used when they could be applied. For feature selection we used the SES algorithm \cite{lagani2017} with $alpha \in \{0.05, 0.01\}$, and $k~\in~\{2, 3\}$ and we also examined the case of no feature selection (i.e., a total of 5 cases/choices). The learning algorithms utilized were Random~Forests \cite{breiman2001}, SVMs \cite{cortes1995}, and LASSO \cite{tibshirani1996}.
For Random~Forests the hyper-parameters and values tried are $numTrees = 1000$, $minLeafSize \in \{1, 3, 5\}$ and $numVarToSample \in \{(0.5, 1, 1.5, 2) * \sqrt{numVar}\}$, where $numVar$ is the number of variables of the dataset. We tested SVMs with linear, polynomial and radial basis function (RBF) kernels. For their hyper-parameters  we examined, wherever applicable, all the combinations of $degree \in \{2, 3\}$, $gamma \in \{0.01, 0.1, 1, 10, 100\}$ and $cost \in \{0.01, 0.1, 1, 10, 100\}$. Finally, LASSO was tested with  $alpha \in \{0.001, 0.5, 1.0\}$ and 10 different values for $lambda$ which were created independently for each dataset using the glmnet library \cite{glmnet}.

We performed tuning and performance estimation of the final model using CVT, TT, NCV, BBC-CV, BCED-CV, and BBC-CV with 10 repeats (BBC-CV\textsuperscript{10}) for each of the $1060$ created sub-datasets, leading to more than 135 million trained models. We set $B = 1000$ for the BBC-CV method, and $B = 1000$, $a = 0.99$ for the BCED-CV method. We applied the same split of the data into $K = 10$ stratified folds for all the protocols. The inner cross-validation loop of NCV uses $K = 9$ folds. It is important to remind at this point that BBC-CV selects the best configuration by estimating the performance of the models on the pooled out-of-sample predictions from all folds. For metrics such as the AUC it is possible that this approach selects a different configuration from the one that the conventional CVT procedure selects (i.e. the configuration with the maximum/minimum average performance/loss over all folds). In anecdotal experiments, we compared the two approaches in terms of the true performance of the models that they return, and found that they perform similarly. In the sections that follow we present the results of CVT with pooling of the out-of-sample predictions. For each protocol, original dataset $D$, and sample size $N$, the results are averaged over the $20$ randomly sampled sub-datasets.

\subsubsection{Bias estimation}

The bias of estimation is computed as in the simulation studies, i.e., $\widehat{Bias} = \hat{P} - P$, where $\hat{P}$ and $P$ denote the estimated and the true performance of the selected configuration, respectively.

In Figure~\ref{fig:real_average_bias} we examine the average bias of the CVT, TT, NCV, BBC-CV, and BCED-CV estimates of performance, on all datasets, relative to sample size. We notice that the results are in agreement with those of the simulation studies. In particular, CVT is optimistically biased for sample size $N \leq 100$ and its bias tends to zero as $N$ increases. TT over-estimates performance for $N = 20$, its bias varies with dataset for $N = 40$, and it over-corrects the bias of CVT for $N \geq 60$. TT exhibits the worst results among all protocols except CVT.

Both NCV and BBC-CV have low bias (in absolute value) regardless of sample size, though results vary with dataset. BBC-CV is mainly conservative with the exception of the \textit{madeline} dataset for $N = 40$ and the \textit{madelon} dataset for $N \in \{60, 80, 100\}$. NCV is slightly optimistic for the \textit{dexter} and \textit{madeline} datasets for $N = 40$ with a bias of 0.033 and 0.031 points of AUC respectively. BCED-CV has, on average, greater bias than BBC-CV for $N \leq 100$. For $N = 500$, its bias shrinks and becomes identical to that of BBC-CV and NCV.

\subsubsection{Relative Performance and Speed Up of BCED-CV}

\begin{figure}[!t]
\centering
\begin{tabular}{c}
\includegraphics[scale=0.65, trim=20 0 0 0]{real_relative_performance_20_small.eps}\\
\includegraphics[scale=0.45, trim=0 0 0 0]{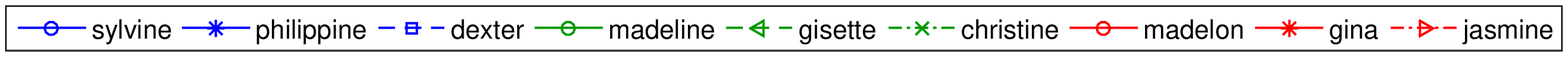}
\end{tabular}
\caption{Relative average true performance of the models returned by the BCED-CV and CVT. For $N \leq 100$ the loss in performance varies greatly with dataset, however, for $N = 500$ there is negligible to no loss in performance. If $N$ if fairly large, BCED-CV will accelerate the CVT procedure without sacrificing the quality of the resulting model or the accuracy of its performance estimate. %except for the \textit{madelon} dataset which exhibits the higher average loss of $1.4\%$.
\label{fig:real_bced_perf}}
\end{figure}

\begin{figure}[!t]
\centering
\begin{tabular}{cc}
\includegraphics[scale=0.65, trim=45 0 0 0]{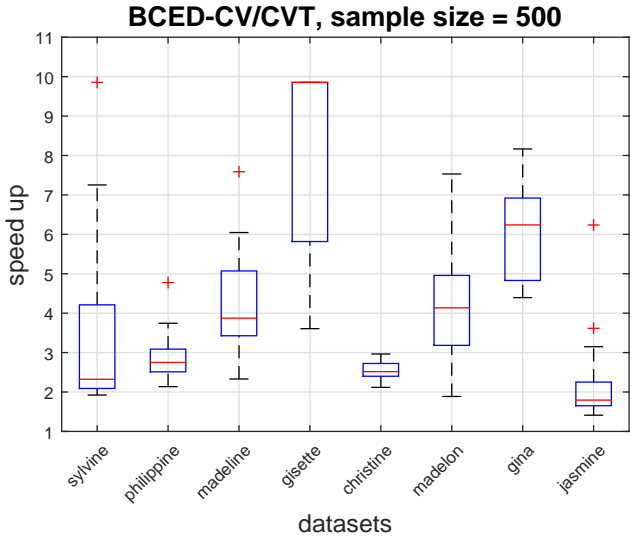}
\end{tabular}
\caption{The speed-up of BCED-CV over CVT is shown for sample size $N = 500$. It is computed as the ratio of models trained by CVT over BCED-CV. Typically, BCED-CV achieves a speed-up of 2-5, up to 10 for the \textit{gisette} dataset. Overall, using BCED-CV results in a significant speed boost, without sacrificing model quality or performance estimation.
\label{fig:real_speed}}
\end{figure}

We have shown that for large sample sizes ($N = 500$) BCED-CV provides accurate estimates of performance of the model it returns, comparable to those of BBC-CV and NCV. How well does this model perform though? In this section, we evaluate the effectiveness of BCED-CV in terms of its tuning (configuration selection) properties, and its efficiency in reducing the computational cost of CVT. %We measure computational cost in terms of the number of models that the protocols train in total. 

Figure~\ref{fig:real_bced_perf} shows the relative average true performance of the models returned by the BCED-CV and CVT protocols, plotted against sample size. We remind here that for each of the 20 sub-datasets of sample size $N \in \{20, 40, 60, 80, 100, 500\}$ sampled from $D_{pool}$, the true performance of the returned model is estimated on the $D_{holdout}$ set. We notice that, for $N \leq 100$ the loss in performance varies greatly with dataset and is quite significant; up to $9.05\%$ in the worst case (\textit{dexter} dataset, $N = 40$). For $N = 500$, however, there is negligible to no loss in performance. Specifically, for the \textit{sylvine, philippine, madeline, christine} and \textit{gina} datasets there is no loss in performance when applying BCED-CV, while there is $0.44\%$ and $0.15\%$ loss for the \textit{gisette} and \textit{jasmine} datasets, respectively. \textit{madelon} exhibits the higher average loss of $1.4\%$. We expect the difference in performance between BCED-CV and CVT to shrink even further with larger sample sizes.

We investigated the reason of the performance loss of BCED-CV for low sample sizes ($N \leq 100$). We observed that, in most cases the majority of configurations ($> 95\%$) were dropped very early within the CV procedure (in the first couple of iterations). With 10-fold CV, the number of out-of-sample predictions with $N \leq 100$ samples ranges from 2 to 10, which are not sufficient for the bootstrap test to reliably identify under-performing configurations. This observation leads to some practical considerations and recommendations. For small sample sizes, we recommend to start dropping configurations with BCED-CV after enough out-of-sample predictions are available. An exact number is hard to determine, as it depends on many factors, such as the analyzed dataset and the set of configurations tested. Given that with $N = 500$ BCED-CV incurs almost no loss in performance, we recommend a minimum of 50 out-of-sample predictions to start dropping configurations, although a smaller number may suffice. For example, with $N = 100$, this would mean that dropping starts after the fifth iteration. Finally, we note that dropping is mostly useful with larger sample sizes (i.e. for computationally costly scenarios), which are also the cases where BCED-CV is on par with BBC-CV and NCV, in terms of tuning and performance estimation.

Next, we compare the computational cost of BCED-CV to CVT, in terms of total number of models trained. The results for $N = 500$ are shown in Figure~\ref{fig:real_speed}. We only focused on the $N = 500$ case, as it is the only case where both protocols produce models of comparable performance. We observe that a speed-up of 2 to 5 is typically achieved by BCED-CV. For the \textit{gisette} dataset, the speed-up is very close the theoretical maximum of this experimental setup. Overall, if sample size is sufficiently large, using dropping is recommended to speed-up CVT without a loss of performance.

\subsubsection{Multiple Repeats}

\begin{figure}[!t]
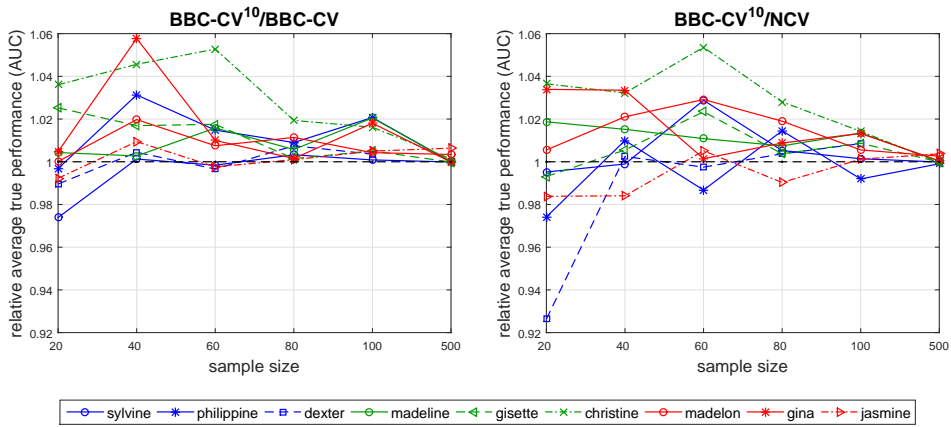

\centering
\begin{tabular}{cc}
\includegraphics[scale=0.45, trim=20 0 0 0]{real_relative_performance_cvt_repeats.eps} & \includegraphics[scale=0.45, trim=0 0 0 0]{real_relative_performance_ncv_repeats.eps}\\
\multicolumn{2}{c}{\includegraphics[scale=0.45, trim=0 0 0 0]{real_legend_h.eps}}
\end{tabular}
\caption{Relative average true performance of BBC-CV\textsuperscript{10} to BBC-CV (left), and of BBC-CV\textsuperscript{10} to NCV (right). Multiple repeats increase the performance of the returned models, maintaining the accuracy of the performance estimation. If computational time is not a limitation, it is preferable to use BBC-CV\textsuperscript{10} over NCV.
\label{fig:real_repeats_perf}}
\end{figure}

We repeated the previous experiments, running BBC-CV with 10 repeats (called BBC-CV\textsuperscript{10} hereafter).
First, we compare the true performance of the models returned by BBC-CV and BBC-CV\textsuperscript{10}, as well as the bias of the estimation. Ideally, using multiple repeats should result in a better performing model, as the variance of the performance estimation (used by CVT for tuning) due to a specific choice of split for the data is reduced when multiple splits are considered. This comes at a cost of increased computational overhead, which in case of 10 repeats is similar to that of the NCV protocol. %To determine and therefore we perform the same comparison against NCV
To determine which of the approaches is preferable, we also compare the performance of the final models produced by BBC-CV\textsuperscript{10} and NCV.

Figure~\ref{fig:real_repeats_perf} (left) shows the relative average true performance of BBC-CV\textsuperscript{10} to BBC-CV with increasing sample size $N$. We notice that, for $N = 20$ the results vary with dataset, however, for $N \geq 40$, BBC-CV\textsuperscript{10} systematically returns an equally well or (in most cases) better performing model than the one that BBC-CV returns. In terms of the bias of the performance estimates of the two methods, we have found them to be similar.   

Similarly, Figure~\ref{fig:real_repeats_perf} (right) shows the comparison between BBC-CV\textsuperscript{10} and NCV. We see again that for sample size $N = 20$ the relative average true performance of the returned models vary with dataset. BBC-CV\textsuperscript{10} outperforms NCV for $N \geq 40$ except for the \textit{philippine} and \textit{jasmine} datasets for which results vary with sample size. Thus, if computational time is not a limiting factor, it is still beneficial to use BBC-CV with multiple repeats instead of NCV.  

To summarize, we have shown that using multiple repeats increases the quality of the resulting models as well as maintaining the accuracy of the performance estimation. We note that the number 10 was chosen mainly to compare BBC-CV to NCV on equal grounds (same number of trained models). If time permits, we recommend using as many repeats as possible, especially for low sample sizes. For larger sample sizes, usually one or a few repeats suffice. 

\subsubsection{Confidence Intervals}

\begin{figure}[!t]
\centering
\begin{tabular}{ccc}
\includegraphics[scale=0.3, trim=45 0 0 0]{real_CIs_true_estimated_ss_20.eps} &
\includegraphics[scale=0.3, trim=30 0 0 0]{real_CIs_true_estimated_ss_100.eps} &
\includegraphics[scale=0.3, trim=30 0 0 0]{real_CIs_true_estimated_ss_500.eps}  \\
\multicolumn{3}{c}{\includegraphics[scale=0.55, trim=0 0 0 0, clip=true]{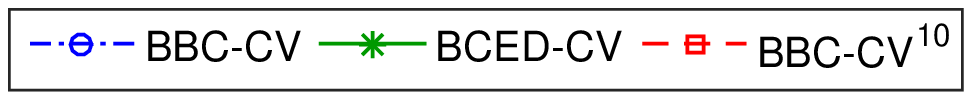}}
\end{tabular}
\caption{Coverage of the $\{50\%, 55\%, \dots, 95\%, 99\%\}$ CIs returned by BBC-CV, BCED-CV, and BBC-CV\textsuperscript{10}, defined as the ratio of the estimated CIs that contain the corresponding true performances of the produced models. The CIs are mainly conservative and become more accurate with increasing sample size and multiple repeats. 
\label{fig:real_cis}}
\end{figure}

The bootstrap-based estimation of performance, allows for easy computation of confidence intervals (CIs) as described in Section~\ref{sec:conf_intervals}. We investigated the accuracy of the CIs produced by the proposed BBC-CV, BCED-CV and BBC-CV\textsuperscript{10} protocols. For this, we computed the coverage of the $\{50\%, 55\%, \dots, 95\%, 99\%\}$ CIs estimated by the protocols, defined as the ratio of the computed CIs that contain the corresponding true performances of the produced models.
%of the true performance of the produced models falling within the respective CI. 
For a given sample size, the coverage of a CI was computed over all 20 sub-datasets and 9 datasets. To further examine the effect of multiple repeats on CIs, we computed their average width (over all 20 sub-datasets) for each dataset and different number of repeats (1 to 10).

\begin{figure}[!t]
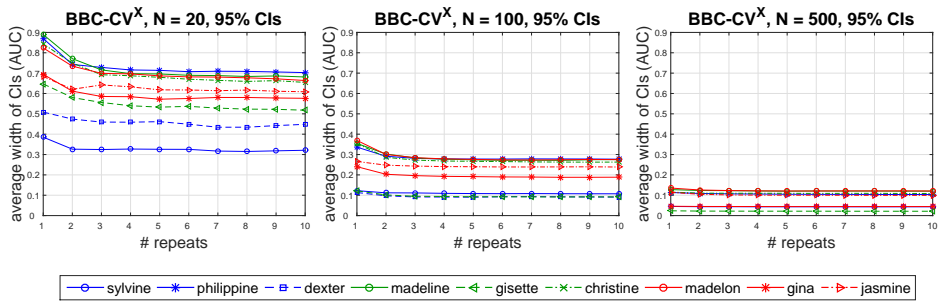

\centering
\begin{tabular}{ccc}
\includegraphics[scale=0.3, trim=45 0 0 0]{real_BBC_repeat_cis_width_20_0_95.eps} &
\includegraphics[scale=0.3, trim=30 0 0 0]{real_BBC_repeat_cis_width_100_0_95.eps} &
\includegraphics[scale=0.3, trim=30 0 0 0]{real_BBC_repeat_cis_width_500_0_95.eps}\\
\multicolumn{3}{c}{\includegraphics[scale=0.45, trim=0 0 0 0]{real_legend_h.eps}}
\end{tabular}
\caption{Average width (over all 20 sub-datasets) of CIs returned by BBC-CV, BCED-CV, and BBC-CV\textsuperscript{10}, for each dataset, with increasing number of repeats. CIs shrink with increasing sample size and number of repeats.
\label{fig:real_cis_width}}
\end{figure}

Figure~\ref{fig:real_cis} shows the estimated coverage of the CIs constructed with the use of the percentile method relative to the expected coverage for the BBC-CV, BCED-CV, and BBC-CV\textsuperscript{10} protocols. We present results for sample sizes $N = 20$ (left), $N = 100$ (middle), and $N = 500$ (right). Figure~\ref{fig:real_cis_width} shows, for the same values for $N$ and for each dataset, the average width of the CIs with increasing number of repeats.

We notice that for $N = 20$ the CIs produced by BBC-CV are conservative, that is, they are wider than ought to be. As sample size increases ($N \geq 100$), BBC-CV returns more calibrated CIs which are still conservative. The use of 10 repeats (BBC-CV\textsuperscript{10}) greatly shrinks the width of the CIs and improves their calibration (i.e., their true coverage is closer to the expected one). The same holds when using dropping of under-performing configurations (BCED-CV). For $N = 500$ the intervals appear to not be conservative. After closer inspection, we saw that this is caused by two datasets (\textit{madeline} and \textit{jasmine}) for which the majority of the true performances are higher than the upper bound of the CI. We note that those datasets are the ones with the highest negative bias (see Figure~\ref{fig:real_average_bias} for $N = 500$), which implicitly causes the CIs to also be biased downwards, thus failing to capture performance estimates above the CI limits.

In conclusion, the proposed BBC-CV method provides mainly conservative CIs of the true performance of the returned models which become more accurate with increasing sample size. %?
The use of multiple repeats improves the calibration of CIs and shrinks their width, for small sample sizes (less than 100). The use of 3-4 repeats seems to suffice and further repeats provide small added value in CI estimation.

%\subsection{Discussion}

\section{Conclusions}
\label{sec:conclusions}

Pooling together the out-of-sample predictions during cross-validation of multiple configurations (i.e., combinations of algorithms and their hyper-parameter values that leads to a model) and employing bootstrapping techniques on them solves in a simple and general way three long-standing, important data analysis tasks: (a) removing the optimism of the performance estimation of the selected configuration, (b) estimating confidence intervals of performance, and (c) dropping from further consideration inferior configurations. While other methods have also been proposed, they lack the simplicity and the generality in applicability in all types of performance metrics. The ideas above are implemented in method BBC-CV tackling points (a) and (b) and BCED-CV that includes (c). 

Simulation studies and experiments on real datasets show empirically that BBC-CV and BCED-CV outperform the alternatives (nested Cross-Validation and the TT method) by either providing more accurate, almost unbiased, conservative estimates of performance even for smaller sample sizes and/or by having much lower computational cost (speed-up of up to 10). We examined the effect of repeatedly applying our methods on multiple fold partitions of the data, and found that we acquire better results in terms of tuning (i.e., better-performing configurations are selected) compared to BBC-CV and NCV. Finally, in our experiments, the confidence intervals produced by bootstrapping are shown to be mainly conservative, improving with increasing sample size and multiple repeats. 

Future work includes a thorough evaluation of the methods on different types of learning tasks such as regression, and survival analysis (however, preliminary results have shown that they are equivalently efficient and effective). 

For a practitioner, based on the results on our methods we offer the following suggestions: first, to forgo the use of the computationally expensive nested cross-validation. Instead, we suggest the use of BBC-CV for small sample sizes (e.g., less than 100 samples). BCED-CV could also be used in these cases to reduce the number of trained models (which may be negligible for such small sample sizes) but it may select a slightly sub-optimal configuration. For larger sample sizes, we advocate the use BCED-CV that is computationally more efficient and maintains all benefits of BBC-CV. We also suggest using as many repeats with different partitions to folds as computational time allows, particularly for small sample sizes, as they reduce the widths of the confidence intervals and lead to a better selection of the optimal configuration.

\begin{acknowledgements}
IT, MT and GB received funding from the European Research Council under the European Union's Seventh Framework Programme (FP/2007-2013) / ERC Grant
Agreement n. 617393. EG received funding from the Toshiba project: ``Feasibility study towards the Next Generation of statistical Text to Speech Synthesis System".
We'd like to thank Pavlos Charoniktakis for constructive feedback.
\end{acknowledgements}

% future work

\newpage

%\begin{acknowledgements}
%\end{acknowledgements}

\bibliographystyle{spmpsci}      % mathematics and physical sciences
\bibliography{vivlio}   % name your BibTeX data base


%\begin{thebibliography}{}
% etc
%\end{thebibliography}

\end{document}